\documentclass[preprint,12pt]{elsarticle}
\usepackage{geometry}
\usepackage{amssymb}
\usepackage{amsmath}
\usepackage{amsthm}
\newtheorem{assumption}{Assumption}

\usepackage{booktabs}
\usepackage{booktabs}
\usepackage{times}
\usepackage{url}
\usepackage{float}
\usepackage{bm}
\usepackage{mathrsfs}

\usepackage{amsmath}
\usepackage{longtable}
\usepackage{amsfonts}
\usepackage{amssymb}

\usepackage{graphicx}
\usepackage{epstopdf}

\usepackage{multirow}
\usepackage{subcaption}
\usepackage{verbatim}
\usepackage{lineno}
\usepackage{enumitem}
\usepackage{color}
\usepackage{subcaption}
\usepackage{epsfig}

\usepackage{lineno}
\modulolinenumbers[5]

\usepackage{bbm} % For indicator function \mathbbm{1}.
\usepackage[T1]{fontenc}
\usepackage{array} 
\usepackage{algorithm}
\usepackage{algpseudocode}
\usepackage{amsmath}

\usepackage{ragged2e}

\usepackage{hyperref}
\usepackage{xcolor}

\journal{}

\begin{document}

\begin{frontmatter}

\title{Simulation of collision avoidance behavior in crowd movement by data-driven approach}

\author[mymainaddress]{Xuanwen Liang}
\author[mymainaddress]{Eric Wai Ming Lee\corref{mycorrespondingauthor}}
\cortext[mycorrespondingauthor]{Corresponding author}
\ead{ericlee@cityu.edu.hk}	
\address[mymainaddress]{Department of Architecture and Civil Engineering\\City University of Hong Kong, Hong Kong}

\begin{abstract}
Crowd movement simulation is essential for pedestrian safety management and facility layout optimization. Data-driven models enhance trajectory prediction accuracy under Euclidean metrics, yet they suffer from excessively high collision rates, especially in bidirectional and multidirectional flows. In this paper, we establish a novel data-driven crowd simulation model that incorporates the pedestrian collision mechanism into the loss function to reduce collisions. A new lateral-acceleration-based collision loss function and a Voronoi-based motion feature extraction approach are proposed. The model is based on a Generative Adversarial Network (GAN) architecture and is termed CPGAN (Collision‑Penalized GAN). We evaluate CPGAN in bidirectional flow scenarios, which involve frequent collision avoidance behaviors. Results show that the proposed lateral-acceleration-based collision loss significantly reduces opposite-direction pedestrian collision rates to levels comparable with controlled experiments. CPGAN effectively simulates bidirectional flow, reproducing lane formation and $N$-$t$ curves. The research outcomes can provide inspiration for integrating pedestrian dynamics mechanisms into loss functions in data-driven crowd simulation.

\end{abstract}


\end{frontmatter}

\section{Introduction}
Crowd simulation is essential for building and pedestrian facility design. Data-driven models based on deep learning have garnered increasing attention for their ability to achieve lower trajectory distance errors than knowledge-driven approaches, such as the social force model \cite{Helbing2000} and cellular automaton model \cite{VARAS2007631}. Specialized data-driven models have been developed for various scenarios, including corridors \cite{ZHAO2020, ZHAO2021}, slopes \cite{10808163}, bottlenecks \cite{10077452}, corners \cite{JIANG2025125706}, T-junctions \cite{9898931, WANG2025130775}, and crosswalks \cite{Mayi2016, Mayi2019}. Models with improved adaptability have also been investigated \cite{Song2021, Liang2024, LIANG2026118481}. For instance, the visual-information-driven model proposed by\cite{LIANG2026118481} is applicable to four scenarios: bottleneck, corridor, corner, and T-junction. These models exhibit low trajectory errors, typically measured by Euclidean distance. \par

Nevertheless, a critical limitation persists: data-driven approaches exhibit excessively high collision rates, especially in high-density, bidirectional, and multidirectional flows. In the trajectory prediction field, collision avoidance has been considered in several studies \cite{korbmacher:hal-03793751, Social_LSTM, Social_GAN, TTC-SLSTM, Collision-Free-LSTM, Yue-Jiangbei1, Yue-Jiangbei2}. For example, \cite{Social_LSTM, Social_GAN, Collision-Free-LSTM} pool neighbor information to mitigate collisions, but collision rates still remain high—over 30\% in unidirectional and 50\% in bidirectional flows \cite{korbmacher:hal-03793751}. Hybrid methods \cite{Yue-Jiangbei1, Yue-Jiangbei2} combine deep networks with the social force model to reduce collisions, yet they rely on knowledge-driven components and their efficacy in high-density environments is uncertain. Alternatively, \cite{TTC-SLSTM} incorporated a time-to-collision (TTC) loss into the Social Long Short-Term Memory (Social-LSTM) model \cite{Social_LSTM} to jointly optimize prediction error and collision risk, yielding improvements over the original model. However, we find it fails to prevent collisions between pedestrians moving in opposite directions. In crowd simulation, several data-driven approaches \cite{Mayi1, Mayi2, ZHAO2020123825, ZHAO2021103260,ZHANG2024110668} modeled bidirectional or multidirectional flows, but they did not explicitly address collisions. Hence, collision handling in data-driven crowd simulation has been largely overlooked.

Collision avoidance is a fundamental pedestrian behavior and has been extensively studied and theorized \cite{PhysRevE.94.022318, QU2021103445, 10.1098/rsos.220187}. This instinctive behavior leads to important self-organization phenomena, such as lane formation in bidirectional flows \cite{Hoogendoorn, FU2022105723, JIN2019137} and stripe formation in crossing flows \cite{Hoogendoorn, stripe2}. These patterns emerge as pedestrians avoid oncoming individuals while following those moving in the same direction, thereby reducing inter-stream interactions and optimizing overall flow efficiency \cite{stripe1}. It is found that pedestrians prefer to change direction rather than reduce speed when avoiding collisions \cite{PhysRevE.94.022318, 10.1371}. \par

A high incidence of collisions in data-driven crowd simulation produces unrealistic behavior and hinder the reproducibility of self-organization phenomena. Effectively reducing collisions is therefore critical. Achieving collision avoidance in data-driven crowd simulation presents a significant challenge, as solely modifying input features or network architectures—a direction that prior work has primarily focused on—proves inadequate \cite{Social_LSTM, Social_GAN, Collision-Free-LSTM}. Although TTC loss \cite{TTC-SLSTM} underperforms in oncoming-pedestrian collision avoidance, we believe this is mainly because TTC is not well aligned with the underlying mechanisms in such scenarios and approaching this problem from the loss function itself is a viable solution. Motivated by these considerations, we attempt to incorporate theories and mechanisms of oncoming-pedestrian collision avoidance into the loss function to reduce collisions and achieve more realistic simulation. \par

To summarize, excessive collisions remain a core challenge in data-driven crowd simulation, and incorporating collision theory into loss functions offers a promising approach to address this issue. In view of this, this paper proposes a novel data-driven model termed CPGAN (Collision‑Penalized Generative Adversarial Network) to reduce collisions. The contributions of this paper can be summarized as follows:
\begin{itemize}
    \item We emphasize the critical importance of collision avoidance in data-driven crowd simulation and highlight incorporating effective collision theory into loss functions as a promising solution. 
    \item We propose a novel model, CPGAN, which integrates a new lateral-acceleration-based collision loss function and a Voronoi-based motion feature extraction approach.
    \item We conduct experiments on bidirectional flow, and the results show that the proposed lateral-acceleration-based collision loss effectively mitigates collisions.
\end{itemize}

\par
The remainder of the paper is structured as follows. Section 2 illustrates the collision avoidance mechanism between pedestrians moving in opposite directions. Section 3 describes the proposed CPGAN model. Section 4 details the experiments and results. Section 5 presents the conclusions.

\section{Collision avoidance} \label{Collision avoidance section}
Collision avoidance with oncoming pedestrians is the defining behavior in bidirectional pedestrian flow \cite{PhysRevE.94.022318, QU2021103445, Hoogendoorn}, yet data-driven models remain plagued by  large numbers of such collisions \cite{korbmacher:hal-03793751}. Our primary objective is to reduce these collisions by incorporating a collision loss function that leverages theories and mechanisms of oncoming-pedestrian collision avoidance. These mechanisms were identified through analysis of a controlled bidirectional flow experiment carried out by Forschungszentrum Jülich (https://ped.fz-juelich.de/da/doku.php?id=corridor4).  \par

\begin{figure*}[h]
    \centering
    \includegraphics[width=0.9\textwidth]{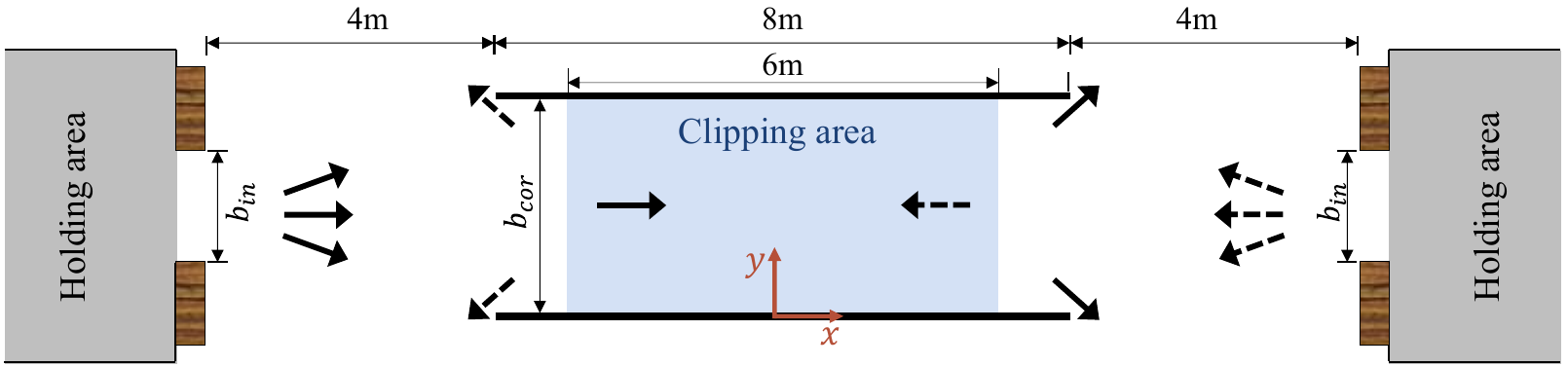}
    \caption{Schematic of the controlled experimental setup (refer to https://ped.fz-juelich.de/da/doku.php?id=corridor4).}
    \label{fig:exp_setup}
    \end{figure*}

\begin{figure*}[h]
    \centering
    \includegraphics[width=0.9\textwidth]{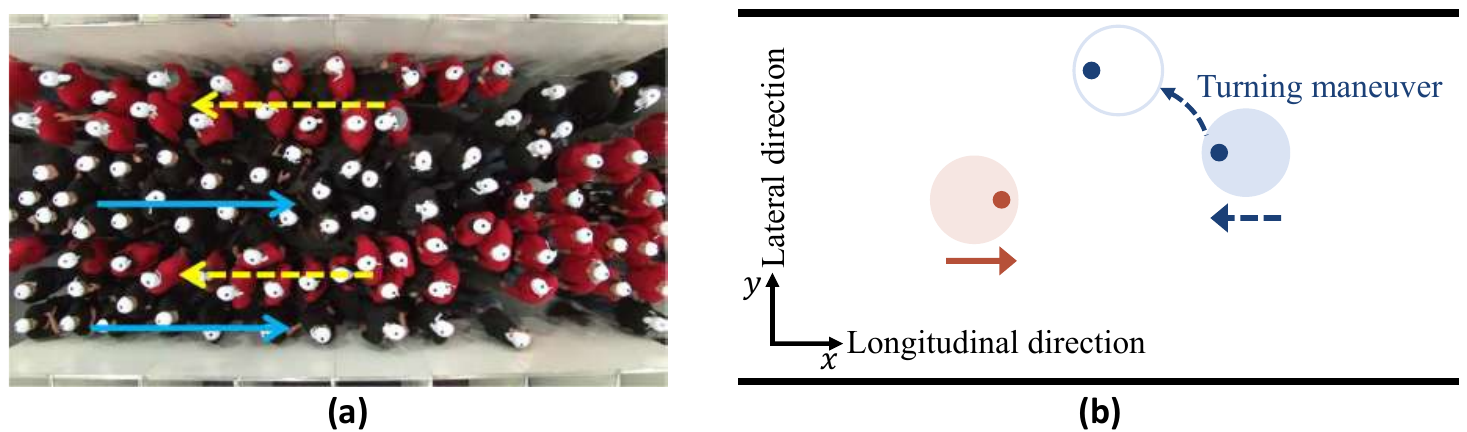}
    \caption{Collision avoidance. (a) Snapshot of the controlled experiment \cite{Zhang_2012}. (b) Dyadic collision-avoidance interaction.}
    \label{fig:collision_avoidance}
    \end{figure*}

Fig. \ref{fig:exp_setup} schematically illustrates the controlled experiment, which was conducted in an 8m‑long corridor. In each run, pedestrians commenced from the holding areas on both sides of the corridor, traverse it, and exit on the opposite side, thereby forming bidirectional flow. Multiple runs were performed by varying the entrance width ($b_{in}$) and the corridor width ($b_{cor}$) to regulate the crowd density. Fig. \ref{fig:collision_avoidance}(a) captures a snapshot of the experiment, where avoidance behavior between oncoming pedestrians can be seen. To investigate the underlying mechanisms, we analyze dyadic (two-pedestrian) collision-avoidance interactions, as shown in Fig. \ref{fig:collision_avoidance}(b). We define the corridor direction ($x$-axis) as the longitudinal direction, and the direction perpendicular to the corridor ($y$-axis) as the lateral direction. Notably, it is observed from the experimental videos that pedestrians exhibit pronounced lateral acceleration to execute a turning maneuver, thereby avoiding the oncoming pedestrian during these interactions. To quantify this, we randomly select 30 dyadic collision-avoidance events and analyze the time evolution of lateral acceleration over the avoidance phase, defined as the interval from the onset of lateral acceleration (maneuver initiation) until the pedestrians align longitudinally (i.e., reach identical $x$-coordinates). \par

\begin{figure*}[h]
    \centering
    \includegraphics[width=0.48\textwidth]{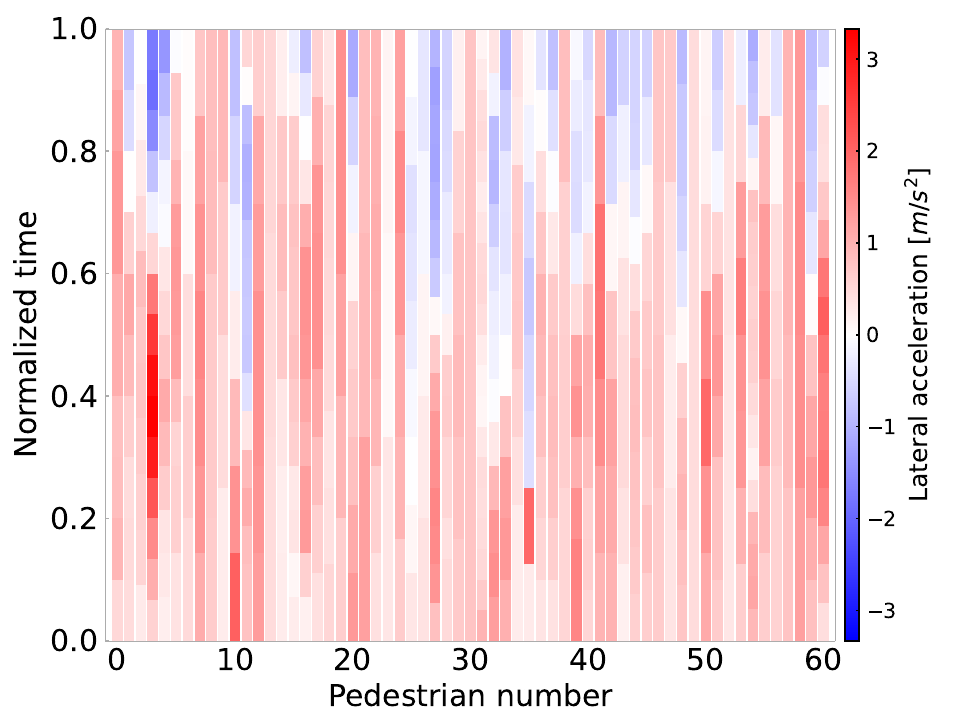}
    \caption{Time evolution of lateral acceleration over the avoidance phase. Here, for each pedestrian, the lateral direction corresponding to their turn is taken as positive.}
    \label{fig:lateral_acc}
    \end{figure*}

Fig. \ref{fig:lateral_acc} displays the time evolution of lateral acceleration over the avoidance phase. For temporal alignment, the phase is normalized to [0,1], with t=0 marking acceleration onset and t=1 marking longitudinal alignment of the two pedestrians. It can be observed that, during the early phase of avoidance, pedestrians typically exhibit a lateral acceleration to turn and avoid collision with the oncoming pedestrian. This lateral acceleration may persist throughout the entire avoidance phase or decrease over time, with the possibility of a reversal in direction. Such a reduction or reversal in lateral acceleration might reflect the pedestrians’ anticipation that the current motion adjustment is sufficient to achieve collision-free interaction, allowing them to prioritize longitudinal progress to reach the corridor exit more efficiently. These findings on collision avoidance enabled by turning (or lateral acceleration) are consistent with prior studies \cite{PhysRevE.94.022318, 10.1371, XIE2022105875, XUAN2026118126}.

\section{Model} \label{sec:model}
\subsection{Overview}
\begin{figure*}[h]
    \centering
    \includegraphics[width=0.9\textwidth]{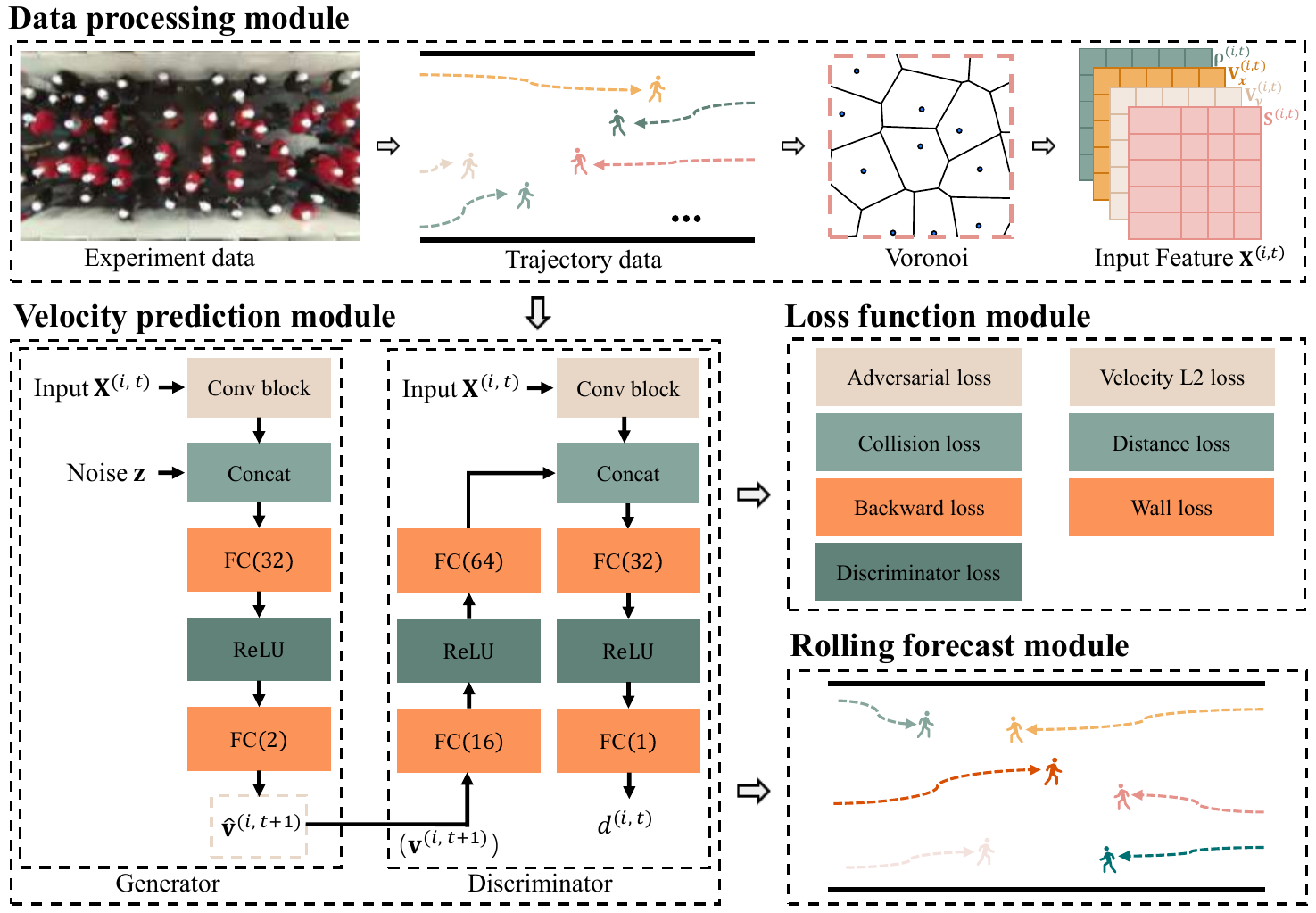}
    \caption{Overview of the CPGAN model.}
    \label{fig:CPGAN_overview}
    \end{figure*}

Considering we are given the current positions and velocities of a group of pedestrians, we aim to simulate the crowd’s future motion. To this end, CPGAN takes the current motion features as input and outputs each pedestrian’s velocity at the next time step. Fig. \ref{fig:CPGAN_overview} presents an overview of CPGAN, which consists of four modules: a data processing (DP) module, a velocity prediction (VP) module, a loss function (LF) module, and a rolling forecast (RF) module. The DP module extracts pedestrian motion features, which are passed to the VP module—a neural network that learns pedestrian motion dynamics and predicts the next-step velocities. The network is trained using the loss functions in the LF module. Once trained, the RF module uses the trained network to perform the crowd simulation.

\subsection{Data processing (DP) module}\label{DP_module}
The DP module is responsible for extracting features that govern pedestrian motion, primarily arising from the pedestrian’s and neighbors’ motion as well as the intended destination \cite{Song2021, ZHAO2021}. To effectively capture these features, we propose a Voronoi-based motion feature extraction approach. Four features are extracted: density $\boldsymbol{\rho}^{(i,t)}$, $x$-velocity $\mathbf{V}_x^{(i,t)}$, $y$-velocity $\mathbf{V}_y^{(i,t)}$, and scene field $\mathbf{S}^{(i,t)}$. The details of the Voronoi-based extraction approaches and these four features are described below.

\subsubsection{Voronoi-based feature extraction}\label{Voronoi-based_feature_extraction}
\begin{figure*}[h]
    \centering
    \includegraphics[width=0.7\textwidth]{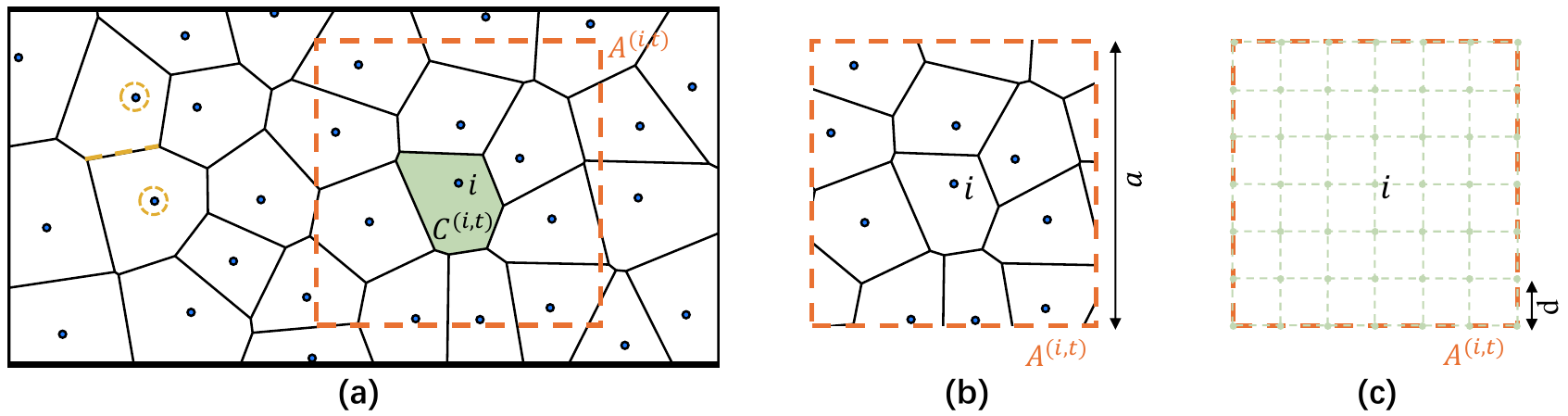}
    \caption{Schematic of the Voronoi-based feature extraction. (a) Voronoi diagram in a corridor at time step $t$. (b) Social interaction area $A^{(i,t)}$ for subject pedestrian $i$ at time step $t$. (c) Discretization of the social interaction area $A^{(i,t)}$.}
    \label{fig:Voronoi}
    \end{figure*}

The Voronoi diagram is widely employed in pedestrian dynamics due to its effectiveness in quantifying local density and velocity distributions \cite{STEFFEN20101902, Zhang_2011}. Moreover, it provides a natural means of identifying neighboring pedestrians. Fig. \ref{fig:Voronoi}(a) illustrates a Voronoi diagram constructed in a corridor, where filled circles denote pedestrians at time step $t$. For each pedestrian $i$, the diagram defines a corresponding Voronoi cell $C^{(i,t)}$ (exemplified by the green polygon in Fig. \ref{fig:Voronoi}(a)), which is the set of all points in the corridor that are closer to pedestrian $i$ than to any other. Two pedestrians are considered Voronoi neighbors if their respective cells share a common edge. For example, the two pedestrians circled with yellow dashed rings in Fig. \ref{fig:Voronoi}(a) are Voronoi neighbors, as their cells share the yellow dashed edge. \par

Following \cite{STEFFEN20101902}, the density $\rho(\mathbf{p}, t)$ at a point $\mathbf{p}$ in the corridor at time step $t$ is then defined as
\begin{equation}
\rho(\mathbf{p}, t) = 
\begin{cases}
\dfrac{1}{|C^{(i,t)}|}, & \text{if } \mathbf{p} \in C^{(i,t)} \\
0, & \text{otherwise}
\end{cases}
\end{equation}
where $|C^{(i,t)}|$ denotes the area of the Voronoi cell $C^{(i,t)}$. Similarly, the velocity components at point $\mathbf{p}$ along the $x$- and $y$-axes, denoted $v_x(\mathbf{p}, t)$ and $v_y(\mathbf{p}, t)$, can be defined as
\begin{equation}
v_x(\mathbf{p}, t) = 
\begin{cases}
v^{(i,t)}_x, & \text{if } \mathbf{p} \in C^{(i,t)} \\
0, & \text{otherwise}
\end{cases}
\end{equation}

\begin{equation}
v_y(\mathbf{p}, t) = 
\begin{cases}
v^{(i,t)}_y, & \text{if } \mathbf{p} \in C^{(i,t)} \\
0, & \text{otherwise}
\end{cases}
\end{equation}
where $v^{(i,t)}_x$ and $v^{(i,t)}_y$ represent the velocity components along the $x$- and $y$-axes of pedestrian $i$ at time step $t$, respectively.

\subsubsection{Features}
Features influencing pedestrian motion include individual movement and social interactions with nearby pedestrians. Therefore, for each subject pedestrian $i$ at time step $t$ ($i, t$), we define a social interaction area $A^{(i,t)}$ to extract these features. $A^{(i,t)}$ is a square of side length $a$ centered on pedestrian $i$, as illustrated by the orange dashed box in Fig. \ref{fig:Voronoi}. To enable neural-network processing, we treat $A^{(i,t)}$ as an image and discretize it into a pixel grid. Specifically, we sample pixel points on $A^{(i,t)}$ at spacing $d$ centered on pedestrian $i$, as illustrated in Fig. \ref{fig:Voronoi}(c). The green dots in Fig. \ref{fig:Voronoi}(c) indicate the sampled pixels. This discretization enables an $\left(\frac{a}{d} + 1\right) \times \left(\frac{a}{d} + 1\right)$ matrix representation of the interaction area $A^{(i,t)}$. Based on this  representation, we extract the following features: $\boldsymbol{\rho}^{(i,t)}$, $\mathbf{V}_x^{(i,t)}$, $\mathbf{V}_y^{(i,t)}$, and $\mathbf{S}^{(i,t)}$.

\begin{enumerate}
    \item \textbf{Density $\boldsymbol{\rho}^{(i,t)}$.} $\boldsymbol{\rho}^{(i,t)}$ encodes the density of pedestrian $i$ and their nearby pedestrians at time step $t$. From the $\left(\frac{a}{d} + 1\right)\times\left(\frac{a}{d} + 1\right)$ sampled pixel points over the interaction area $A^{(i,t)}$, we compute the density at each point (following the method in Section \ref{Voronoi-based_feature_extraction}) and form the matrix $\boldsymbol{\rho}^{(i,t)}$. Thus, $\boldsymbol{\rho}^{(i,t)} \in \mathbb{R}^{\left(\frac{a}{d} + 1\right)\times\left(\frac{a}{d} + 1\right)}$.

    \item \textbf{Velocity along the $x$-axis, $\mathbf{V}_x^{(i,t)}$.} $\mathbf{V}_x^{(i,t)}$ represents the $x$-velocity of pedestrian $i$ and their nearby pedestrians at time step $t$. Similarly, $\mathbf{V}_x^{(i,t)} \in \mathbb{R}^{\left(\frac{a}{d} + 1\right)\times\left(\frac{a}{d} + 1\right)}$ comprises the $x$-velocity values at each pixel point.

    \item \textbf{Velocity along the $y$-axis, $\mathbf{V}_y^{(i,t)}$.} $\mathbf{V}_y^{(i,t)} \in \mathbb{R}^{\left(\frac{a}{d} + 1\right)\times\left(\frac{a}{d} + 1\right)}$ encodes the $y$-velocity at each pixel point.

    \item \textbf{Scene field $\mathbf{S}^{(i,t)}$.} Analogous to the static field in cellular automaton models \cite{VARAS2007631}, $\mathbf{S}^{(i,t)}$ incorporates the influence of the destination. Specifically, $\mathbf{S}^{(i,t)} \in \mathbb{R}^{\left(\frac{a}{d} + 1\right)\times\left(\frac{a}{d} + 1\right)}$, where each element stores the distance from a pixel point to the destination.
\end{enumerate}
\par

Lastly, we stack the four two-dimensional matrices to form $\mathbf{X}^{(i,t)}$:
\[
\mathbf{X}^{(i,t)} = \text{stack}( \boldsymbol{\rho}^{(i,t)}, \mathbf{V}_x^{(i,t)}, \mathbf{V}_y^{(i,t)}, \mathbf{S}^{(i,t)}) \in \mathbb{R}^{4 \times \left(\frac{a}{d} + 1\right)\times\left(\frac{a}{d} + 1\right)} 
\]

\subsection{Velocity prediction (VP) module}

The VP module generates the velocity of pedestrian $i$ at the next time step $t+1$, ${\hat{\mathbf{v}}^{(i, t+1)}} = [\hat{v}_{x}^{(i, t+1)}, \hat{v}_y^{(i, t+1)}] \in \mathbb{R}^{1 \times 2}$, based on input feature $\mathbf{X}^{(i,t)}$. Here, $\hat{v}_{x}^{(i, t+1)}$ and $\hat{v}_y^{(i, t+1)}$ denote the predicted velocity components along the $x$- and $y$-axes, respectively. We employ a Generative Adversarial Network (GAN) \cite{2014gan} framework to solve this task, since its adversarial training mechanism can effectively capture the multi-modal distribution of possible future motions and enables the generation of diverse, socially plausible, and physically realistic velocity prediction \cite{Social_GAN}. 
\par

\begin{figure*}[h]
    \centering
    \includegraphics[width=0.9\textwidth]{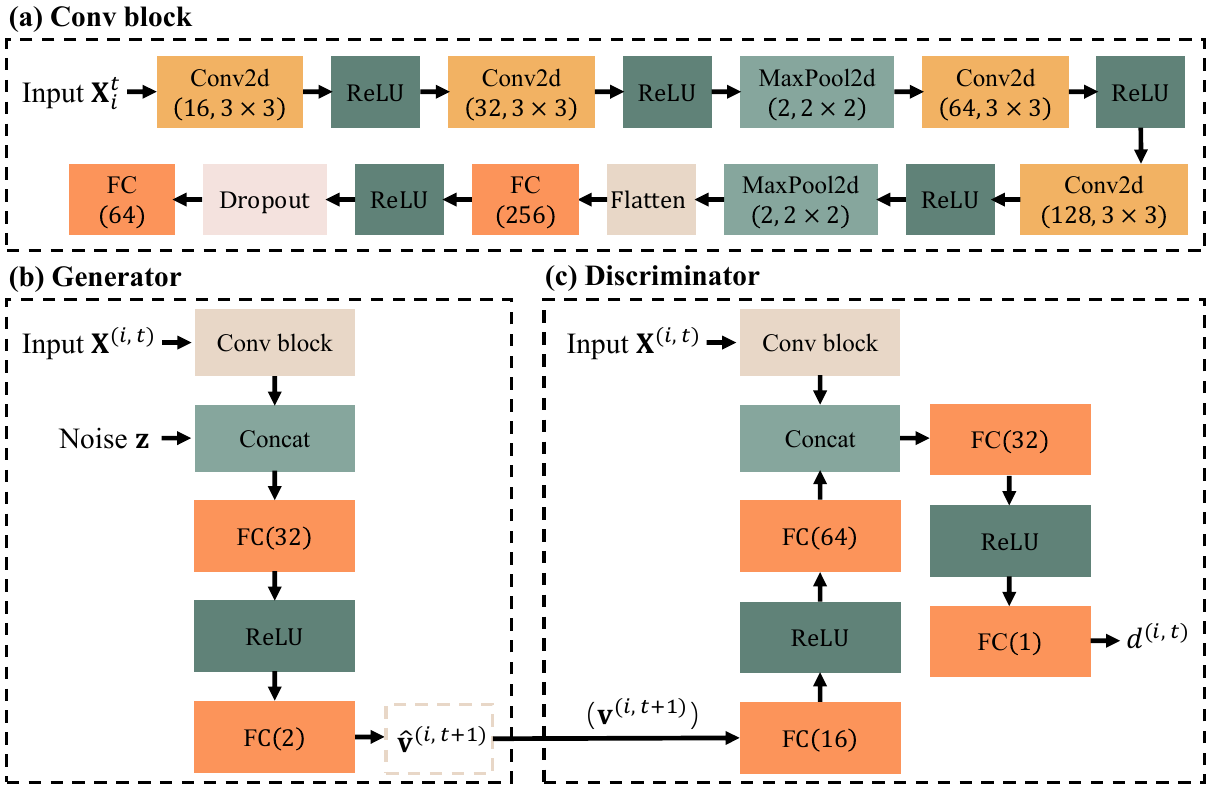}
    \caption{Structure of the VP module. (a) Architecture of the convolutional block. Conv2d ($16, 3 \times 3$) denotes a 2D convolutional layer with 16 output channels and a $3 \times 3$ kernel. MaxPool2d ($2, 2 \times 2$) represents a $2 \times 2$ max-pooling layer with stride 2. FC (256) indicates a fully connected layer with 256 output units. (b) Architecture of the generator. (c) Architecture of the discriminator.}
    \label{fig:VP_module}
    \end{figure*}

Fig. \ref{fig:VP_module} presents the structure of the VP module. As a GAN-based framework, it consists of two components: a generator G (Fig. \ref{fig:VP_module} (b)) and a discriminator D (Fig. \ref{fig:VP_module} (c)). G takes $\mathbf{X}^{(i,t)}$ as input and outputs predicted velocity $\hat{\mathbf{v}}^{(i, t+1)}$. D takes as input $\mathbf{X}^{(i,t)}$ along with either the predicted velocity $\hat{\mathbf{v}}^{(i, t+1)}$ (from G) or the ground-truth velocity $\mathbf{v}^{(i, t+1)}$. It then outputs a discriminative score $d^{(i,t)}$, where a higher value indicates that the input velocity is more likely to be real. Through this architecture, G and D are trained adversarially. The discriminator aims to maximize its ability to distinguish ground-truth velocities from those generated by G. Meanwhile, the generator learns to produce $\hat{\mathbf{v}}^{(i, t+1)}$ that are sufficiently realistic to “fool” D into assigning high scores, thereby driving G to learn plausible and realistic future pedestrian motions. 

\textbf{Generator.} The architecture of the generator G is depicted in Fig. \ref{fig:VP_module} (b). First, the input feature $\mathbf{X}^{(i,t)}$ is encoded by a convolutional (Conv) block to obtain a feature vector $\mathbf{e}^{(i,t)}$. The detailed structure of this Conv block is shown in Fig. \ref{fig:VP_module} (a). It consists of a series of operations, including 2D convolution, rectified linear unit (ReLU) activation \cite{nair2010rectified}, max-pooling, flattening, dropout \cite{srivastava2014dropout}, fully connected (FC) layers. This encoding can be formulated as:
\begin{equation}
\mathbf{e}^{(i,t)} = f_{\text{enc}}(\mathbf{X}^{(i,t)})
\end{equation}
where $f_{\text{enc}}(\cdot)$ denotes the composite function of the sequential operations in the Conv block. Unlike traditional GANs in which G takes random noise as input to generate unconstrained samples \cite{2014gan}, our aim is to generate future motions that are consistent with the observed current state. To this end, we introduce conditioning by concatenating the encoded current feature $\mathbf{e}^{(i,t)}$ with a random noise vector $\mathbf{z} \sim \mathcal{N}(0, 1)$:
\begin{equation}
\mathbf{h}_G^{(i,t)}=\left[\mathbf{e}^{(i,t)}; \mathbf{z}\right]
\end{equation}
The concatenated representation $\mathbf{h}_G^{(i,t)}$ is then used to synthesize the predicted velocity $\hat{\mathbf{v}}^{(i, t+1)}$:
\begin{equation}
\hat{\mathbf{v}}^{(i, t+1)} = \phi(\mathbf{h}_G^{(i,t)})
\end{equation}
where $\phi(\cdot)$ consists of two fully connected layers with a ReLU activation in between.

\par
\textbf{Discriminator.} The architecture of the discriminator D is shown in Fig. \ref{fig:VP_module} (c). Similar to G, the input $\mathbf{X}^{(i,t)}$ is first encoded by the Conv block to obtain a contextual feature representation. Meanwhile, the velocity candidate $\hat{\mathbf{v}}^{(i, t+1)}$ (which can be either the generated velocity $\hat{\mathbf{v}}^{(i, t+1)}$ from G or the ground truth $\mathbf{v}^{(i, t+1)}$) is separately processed by a sequence of layers. The outputs from these two parallel branches are then concatenated, yielding
\begin{equation}
\mathbf{h}_D^{(i,t)}=\left[f_{\text{enc}}(\mathbf{X}^{(i,t)}); f_{\text{vel}}(\mathbf{v}^{(i, t+1)})\right]
\end{equation}
where $f_{\text{vel}}(\cdot)$ denotes the velocity-processing branch, consisting of two fully connected layers with a ReLU activation. The concatenated representation $\mathbf{h}_D^{(i,t)}$ is subsequently fed into additional layers to produce the final discrimination score $d^{(i,t)}$:
\begin{equation}
d^{(i,t)} = \psi(\mathbf{h}_D^{(i,t)})
\end{equation}
where $\psi(\cdot)$ comprises two fully connected layers with a ReLU activation.

\subsection{Loss function (LF) module}\label{LF_module}
The LF module comprises a series of loss functions used to train the generator G and discriminator D. These loss functions are designed to encourage G to generate more realistic velocity predictions and to enhance D’s ability to discriminate between real and generated velocities.
\subsubsection{Generator losses}
\begin{enumerate}
    \item \textbf{Adversarial loss $\mathcal{L}_{adv}^{(i,t)}$.} $\mathcal{L}_{adv}^{(i,t)}$ encourages the generator to produce velocities $\hat{\mathbf{v}}^{(i, t+1)}$ that are indistinguishable from real data to the discriminator $D$. It is defined as
    \begin{equation}
    \mathcal{L}_{adv}^{(i,t)} = \text{BCEWithLogitsLoss}\Bigl(D\bigl(\mathbf{X}^{(i,t)},\, \hat{\mathbf{v}}^{(i, t+1)}\bigr),\; 1\Bigr)
    \end{equation}
    where the discriminator $D$ takes feature $\mathbf{X}^{(i,t)}$ and the generated velocity $\hat{\mathbf{v}}^{(i, t+1)}$ as input and outputs a discrimination score. The target label 1 corresponds to real data. $\text{BCEWithLogitsLoss}$ denotes the binary cross-entropy loss combined with a sigmoid activation.

    \item \textbf{Velocity L2 loss $\mathcal{L}_{vel}^{(i,t)}$.} $\mathcal{L}_{vel}^{(i,t)}$ measures the squared Euclidean distance between the generated velocity $\hat{\mathbf{v}}^{(i, t+1)}$ and the ground truth $\mathbf{v}^{(i, t+1)}$:
    \begin{equation}
    \mathcal{L}_{vel}^{(i,t)} = \left\| \mathbf{v}^{(i, t+1)} - \hat{\mathbf{v}}^{(i, t+1)} \right\|_2^2
    \end{equation}

    \item \textbf{Collision loss $\mathcal{L}_{col}^{(i,t)}$.} Collision avoidance with oncoming pedestrians is the defining behavior in bidirectional pedestrian flow, facilitating lane formation and improving overall efficiency \cite{PhysRevE.94.022318, QU2021103445, Hoogendoorn}. However, data-driven models exhibit frequent oncoming-pedestrian collisions in bidirectional flow \cite{korbmacher:hal-03793751}. Our primary goal is to reduce such collisions by incorporating a collision loss function that leverages theories and mechanisms of oncoming-pedestrian collision avoidance. According to the mechanisms in Section \ref{Collision avoidance section}, we propose a lateral-acceleration-based collision loss.

    First, compute the acceleration component of pedestrian $i$ in the lateral direction ($y$-axis), $acc^{(i,t)}_y$:
    \begin{equation}
    acc^{(i,t)}_y = (\hat{v}_y^{(i, t+1)} - v^{(i,t)}_y) / (1/ \mathrm{fps})
    \end{equation}
    where $v^{(i,t)}_y$ denotes the $y$-component of pedestrian $i$’s velocity at time step $t$, \(\mathrm{fps}\) is the frame rate (frames per second). Let $\mathcal{V}(i,t)$ denote the set of Voronoi neighbors of $(i,t)$. The set of Voronoi neighbors moving in the opposite direction for $(i,t)$, $\mathcal{N}_{\mathrm{vor}}^{\mathrm{opp}}(i,t)$, is then defined as
    \begin{equation}
    \mathcal{N}_{\mathrm{vor}}^{\mathrm{opp}}(i,t) = \Bigl\{ j \in \mathcal{V}(i,t) \ \Bigl|\ 
    v^{(i,t)}_x \cdot v^{(j,t)}_x < 0 \Bigr\}
    \end{equation}

where $v^{(i,t)}_x$ and $v^{(j,t)}_x$ are the $x$-component velocities of pedestrians $i$ and $j$ at time step $t$, respectively. Since pedestrian $i$ can only control its own collision-avoidance maneuver by executing lateral acceleration and cannot control or anticipate oncoming pedestrians' collision-avoidance actions, we make the following assumption:
\begin{assumption}\label{ass:main}
    Pedestrian $i$ maintains its current lateral acceleration (i.e., $y$-acceleration, $acc^{(i,t)}_y$) and longitudinal velocity (i.e., $x$-velocity), and pedestrian $j$ maintains its current velocity $\mathbf{v}^{(j,t)}$ from time step $t$ onward.
\end{assumption}
 In other words, let the current discrete time step $t$ correspond to the actual continuous time $\tau_0 = t/ \mathrm{fps}$ (in seconds). Then, the $x$- and $y$-velocities at any future time $\tau_0 + \Delta \tau$ for pedestrian $i$ and $j$ are computed as
    \begin{equation}
        v^{(i, \tau_0 + \Delta \tau)}_x = v^{(i, \tau_0)}_x = v^{(i,t)}_x
    \end{equation}
    \begin{equation}
        v^{(i, \tau_0 + \Delta \tau)}_y = v^{(i, \tau_0)}_y +  acc^{(i,t)}_y \times \Delta \tau = v^{(i,t)}_y +  acc^{(i,t)}_y \times \Delta \tau
    \end{equation}
    \begin{equation}
        v^{(j, \tau_0 + \Delta \tau)}_x = v^{(j, \tau_0)}_x = v^{(j,t)}_x
    \end{equation}
    \begin{equation}
        v^{(j, \tau_0 + \Delta \tau)}_y = v^{(j, \tau_0)}_y = v^{(j,t)}_y
    \end{equation}

where $v^{(i,t)}_x$, $v^{(i,t)}_y$, $v^{(j,t)}_x$, $v^{(j,t)}_y$ are the $x$- and $y$-velocities of pedestrians $i$ and $j$, respectively, at time step $t$. Let $\mathbf{p}^{(i,t)} = [p^{(i,t)}_x, p^{(i,t)}_y] \in \mathbb{R}^{1 \times 2}$ denote pedestrian \(i\)'s position (its \(x\)- and \(y\)-coordinates) and $\mathbf{v}^{(i,t)}  = [v^{(i,t)}_x, v^{(i,t)}_y] \in \mathbb{R}^{1 \times 2}$ denote pedestrian \(i\)'s velocity at time step \(t\). Similarly, \(\mathbf{p}^{(j,t)}\) and \(\mathbf{v}^{(j,t)}\) are defined for pedestrian \(j\). Based on Assumption \ref{ass:main}, the expected minimum distance between pedestrians $i$ and $j$ over the time interval $[\tau_0, +\infty)$, $d^{(i,j,t)}_{\min}$, can be calculated as 

    \begin{equation}
    d^{(i,j,t)}_{\min} = f(\mathbf{p}^{(i,t)}, \mathbf{v}^{(i,t)}, acc^{(i,t)}_y, \mathbf{p}^{(j,t)}, \mathbf{v}^{(j,t)})
    \end{equation}
    where $f(\cdot)$ is a function that computes $d^{(i,j,t)}_{\min}$ given $\mathbf{p}^{(i,t)}$, $\mathbf{v}^{(i,t)}$, $acc^{(i,t)}_y$, $\mathbf{p}^{(j,t)}$, $\mathbf{v}^{(j,t)}$. The detailed calculation procedure for $d^{(i,j,t)}_{\min}$ is provided in \ref{ap:Minimum_distance_calculation}. Let $r_{ped}$ denote the pedestrian radius. Here, a collision is considered to occur if the distance between pedestrians $i$ and $j$ is less than $2 \times r_{ped}$. The collision loss for $(i,t)$ is defined as

    \begin{equation} \label{eq:col_loss}
    \mathcal{L}_{col}^{(i,t)} = \sum_{j \in \mathcal{N}_{\mathrm{vor}}^{\mathrm{opp}}(i,t)} \text{ReLU}(2 \times r_{ped} - d^{(i,j,t)}_{\min})
    \end{equation}
    Eq. (\ref{eq:col_loss}) indicates that if the expected minimum distance $d^{(i,j,t)}_{\min} < 2 \times r_{ped}$ (i.e., a collision has occurred at $\tau_0$ or is expected to occur in $(\tau_0, +\infty)$ based on Assumption \ref{ass:main}), the subject pedestrian $i$ is assigned a loss value of $2 \times r_{ped} - d^{(i,j,t)}_{\min}$, which increases as $d^{(i,j,t)}_{\min}$ decreases; otherwise, the loss equals zero.

    \item \textbf{Distance loss $\mathcal{L}_{dis}^{(i,t)}$.} When the distance between pedestrians moving in the same direction is too small, overlapping occurs. Distance loss $\mathcal{L}_{dis}^{(i,t)}$ is introduced to penalize such overlapping. First, the set of Voronoi neighbors moving in the same direction for $(i,t)$ is defined as
    \begin{equation}
    \mathcal{N}_{\mathrm{vor}}^{\mathrm{same}}(i,t) = \Bigl\{ j \in \mathcal{V}(i,t) \ \Bigl|\ 
    v^{(i,t)}_x \cdot v^{(j,t)}_x \geq 0 \Bigr\}
    \end{equation} 
    For neighbor $j \in \mathcal{N}_{\mathrm{vor}}^{\mathrm{same}}(i,t)$, assume that pedestrian $i$ moves from its current position $\mathbf{p}^{(i,t)}$ with the predicted velocity $\hat{\mathbf{v}}^{(i,t+1)}$, while pedestrian $j$ continues moving with its current velocity $\mathbf{v}^{(j,t)}$. Then, the expected distance between pedestrians $i$ and $j$ at time step $t+1$, $d^{(i,j,t+1)}$, is calculated as

    \begin{equation}
    d^{(i,j,t+1)} = \|(\mathbf{p}^{(i,t)} + \hat{\mathbf{v}}^{(i, t+1)} / \mathrm{fps}) - (\mathbf{p}^{(j,t)} + \mathbf{v}^{(j,t)}/ \mathrm{fps})\|
    \end{equation}

    The distance loss for $(i,t)$ is defined as
    \begin{equation}
    \mathcal{L}_{dis}^{(i,t)} = \sum_{j \in \mathcal{N}_{\mathrm{vor}}^{\mathrm{same}}(i,t)} \text{ReLU}(2 \times r_{ped} - d^{(i,j,t+1)})
    \end{equation}

    \item \textbf{Backward loss $\mathcal{L}_{bac}^{(i,t)}$.} Unrealistic pedestrian backward motion can occur in data-driven approaches. Prior work has addressed this issue by either directly reversing the direction of the predicted velocity or constraining pedestrians to move straight toward the target for backward motion \cite{ZHAO2021103260, LIANG2026118481}. Although these strategies are simple, they are rather coarse and lack physical interpretability. To mitigate this limitation, we introduce a novel backward loss function: 
    \begin{equation}
    \mathcal{L}_{bac}^{(i,t)} = \sigma \bigl( - \alpha ( \hat{\mathbf{v}}^{(i, t+1)} \cdot \hat{\mathbf{v}}_d ) \bigr)
    \end{equation}
    where $\sigma(\cdot)$ is the sigmoid function, $\alpha$ is a temperature parameter controlling the steepness of the sigmoid, and $\hat{\mathbf{v}}_d$ is the unit vector in the desired velocity direction.
    
    \item \textbf{Wall loss $\mathcal{L}_{wal}^{(i,t)}$.} Similar to backward motion, wall-crossing behavior is also a common limitation of data-driven models. Previous work has either disregarded this issue or relied on similarly coarse solutions \cite{ZHAO2021103260, LIANG2026118481, Song2021}. We introduce a wall loss function to address this. Specifically, let \(\hat{\mathbf{n}}\) denote the unit vector normal to the wall. The predicted normal displacement of pedestrian \(i\) at time step \(t+1\), $\Delta x_n^{(i, t+1)}$, is computed as
    \begin{equation}
    \Delta x_n^{(i, t+1)} = ( \hat{\mathbf{v}}^{(i, t+1)} \cdot \hat{\mathbf{n}} ) / \mathrm{fps}
    \end{equation}
    Let \(d^{(i,t)}\) denote the distance from pedestrian \(i\) to the wall at time \(t\). Then, $\mathcal{L}_{wal}^{(i,t)}$ is defined as
    \begin{equation}
    Y = \sigma\Bigl(\beta \cdot \bigl(\Delta x_n^{(i, t+1)} - [d^{(i,t)} - (r_{ped} + \delta)]\bigr)\Bigr)
    \end{equation}
    where $\beta$ is a temperature parameter, $r_{ped}$ is the pedestrian radius, and $\delta$ is a small buffer distance.

\end{enumerate}
The total generator loss $\mathcal{L}_{G}^{(i,t)}$ is defined as the sum of these loss components:
    \begin{equation}
    \mathcal{L}_{G}^{(i,t)} = \mathcal{L}_{adv}^{(i,t)} + \mathcal{L}_{vel}^{(i,t)} + \mathcal{L}_{col}^{(i,t)} + \mathcal{L}_{dis}^{(i,t)} + \mathcal{L}_{bac}^{(i,t)} + \mathcal{L}_{wal}^{(i,t)}
    \end{equation}
The generator G's parameters are updated by minimizing the mini-batch average of the total generator loss, $\frac{1}{|\mathcal{B}|} \sum_{(i,t) \in \mathcal{B}} \mathcal{L}_{G}^{(i,t)}$, where $\mathcal{B}$ denotes a mini-batch and $|\mathcal{B}|$ is the batch size.

\subsubsection{Discriminator loss}
The discriminator loss $\mathcal{L}_{D}^{(i,t)}$ aims to enhance D’s ability to discriminate between real and generated velocities. It is defined as
    \begin{equation}
    \mathcal{L}_{D}^{(i,t)} = \text{BCEWithLogitsLoss}\Bigl(D\bigl(\mathbf{X}^{(i,t)},\, \hat{\mathbf{v}}^{(i, t+1)}\bigr),\; 0\Bigr) + \text{BCEWithLogitsLoss}\Bigl(D\bigl(\mathbf{X}^{(i,t)},\, {\mathbf{v}}^{(i, t+1)}\bigr),\; 1\Bigr)
    \end{equation}
where the target labels 0 and 1 correspond to generated and real data, respectively. The discriminator D's parameters are updated by minimizing the mini-batch average of the discriminator loss, $\frac{1}{|\mathcal{B}|} \sum_{(i,t) \in \mathcal{B}} \mathcal{L}_{D}^{(i,t)}$.
    
\subsection{Rolling forecast (RF) module}\label{RF_module}
The RF module simulates continuous crowd movement once the network (the VP module) has been trained. The simulation scenarios are configured identically to the controlled testing experiments (i.e., the testing scenarios listed in Table~\ref{tab:dataset}; see also Section~\ref{sec:dataset} for further details). Specifically, the entrance width ($b_{in}$), corridor width ($b_{cor}$), number of pedestrians, as well as each pedestrian’s entry time and initial position are kept the same as in the testing scenarios to ensure comparability. The RF simulation proceeds iteratively in discrete time steps until all pedestrians have exited. The core loop for each time step $t$ is as follows:
\begin{enumerate}
    \item Entry update: For any pedestrian whose entry time equals the current time step $t$, an agent is instantiated at its corresponding initial position.

    \item Feature extraction: Extract the motion feature $\mathbf{X}^{(i,t)}$ for each pedestrian $i$ present in the scene.

    \item Velocity prediction: The trained network predict each pedestrian $i$'s velocity $\hat{\mathbf{v}}^{(i, t+1)}$ according to their motion feature $\mathbf{X}^{(i,t)}$.

    \item Position update: Update each pedestrian $i$'s position $\mathbf{p}^{(i,t+1)}$ based on its predicted velocity $\hat{\mathbf{v}}^{(i, t+1)}$ and current position $\mathbf{p}^{(i,t)}$, i.e., $\mathbf{p}^{(i,t+1)} = \mathbf{p}^{(i,t)} + \hat{\mathbf{v}}^{(i, t+1)}/\mathrm{fps}$.

    \item Exit removal: Pedestrians whose updated positions have moved beyond the designated exit boundary are removed from the simulation.
\end{enumerate}
The loop terminates once all pedestrians have exited the scene, completing the trajectory simulation for the entire crowd.

\section{Experiments and results}
\subsection{Experiments}
\subsubsection{Datasets} \label{sec:dataset}
Bidirectional flow involves extensive collision avoidance behaviors. Therefore, we use the controlled bidirectional experiments conducted by Forschungszentrum Jülich to construct our dataset. The experimental setup is described in Section \ref{Collision avoidance section}. To enhance the dataset quality, only trajectory data within a clipping area (as shown in Fig. \ref{fig:exp_setup}) with a length of 6m are utilized. Multiple runs were performed by varying the entrance width $\left(b_{in}\right)$ and the corridor width $\left(b_{cor}\right)$ to regulate the crowd density. These runs are divided into two categories: training-validation scenarios and testing scenarios, as illustrated in Tables \ref{tab:dataset}. According to the method described in Section \ref{sec:model}, we extracted the input features and output vectors from the training-validation scenarios to construct the training-validation dataset. This dataset is split in a 4:1 ratio to train and validate the model. After training, we conducted crowd simulations based on the testing scenarios with the rolling forecast module. 

\begin{table*}[htb]
    \centering
    \footnotesize
    \caption{Parameters of the dataset. $N$ denotes the number of pedestrians.}
    \resizebox{0.7\textwidth}{!}{
    \begin{tabular}{
    >{\centering}m{0.2\textwidth}  
    >{\centering}m{0.2\textwidth} >{\centering}m{0.1\textwidth} 
    >{\centering}m{0.1\textwidth} >{\centering\arraybackslash}m{0.1\textwidth}}
      \toprule %Add a bold line to the header
      Dataset  & Name & $b_{in}[m]$ & $b_{cor}[m]$ & $N[Pers.]$ \\ 
      \hline
     \multirow{5}{*}{Training-validation} & 
                          B-300-050 & 3.00 & 0.50 & 125 \\
                         \cline{2-5}
                          & B-300-065 & 3.00 & 0.65 & 147 \\
                         \cline{2-5}
                          & B-300-075 & 3.00 & 0.75 & 147 \\
                         \cline{2-5}
                          & B-300-085 & 3.00 & 0.85 & 216 \\
                         \cline{2-5}
                          & B-300-100 & 3.00 & 1.00 & 230 \\

    \hline
         \multirow{4}{*}{Testing} & 
                          B-360-050 & 3.60 & 0.50 & 130 \\
                         \cline{2-5}
                          & B-360-075 & 3.60 & 0.75 & 127 \\
                         \cline{2-5}
                          & B-360-090 & 3.60 & 0.90 & 212 \\
                         \cline{2-5}
                          & B-360-120 & 3.60 & 1.20 & 309 \\

    \bottomrule
    \end{tabular}}
    \label{tab:dataset}
\end{table*}

\subsubsection{Parameter Settings}
We set the side length $a$ of the social interaction area $A^{(i,t)}$ to 2.4m according to \cite{doi:10.1086/200975}. The spacing $d$ for $A^{(i,t)}$ is set to 0.05m to balance data granularity and dataset size. The temperature parameters $\alpha$ and $\beta$ are set to $100$ to approximate the step function. The pedestrian radius $r_{ped}$ is set to 0.2m. The buffer distance $\delta$ is set to 0.1m considering the pedestrians' physical size. The unit vector in the desired-velocity direction, $\hat{\mathbf{v}}_d$, is set to $[1,0]$ for pedestrians walking to the right and to $[-1,0]$ for pedestrians walking to the left. We train the model for 2000 epochs. Consistent with \cite{2014gan}, we employ the Adam optimizer \cite{kingma2017adam} for both the generator and discriminator, using a learning rate of 0.001. 

\subsection{Results}
\subsubsection{Trajectories and spatial distributions}
\begin{figure*}[h]
    \centering
    \includegraphics[width=0.98\textwidth]{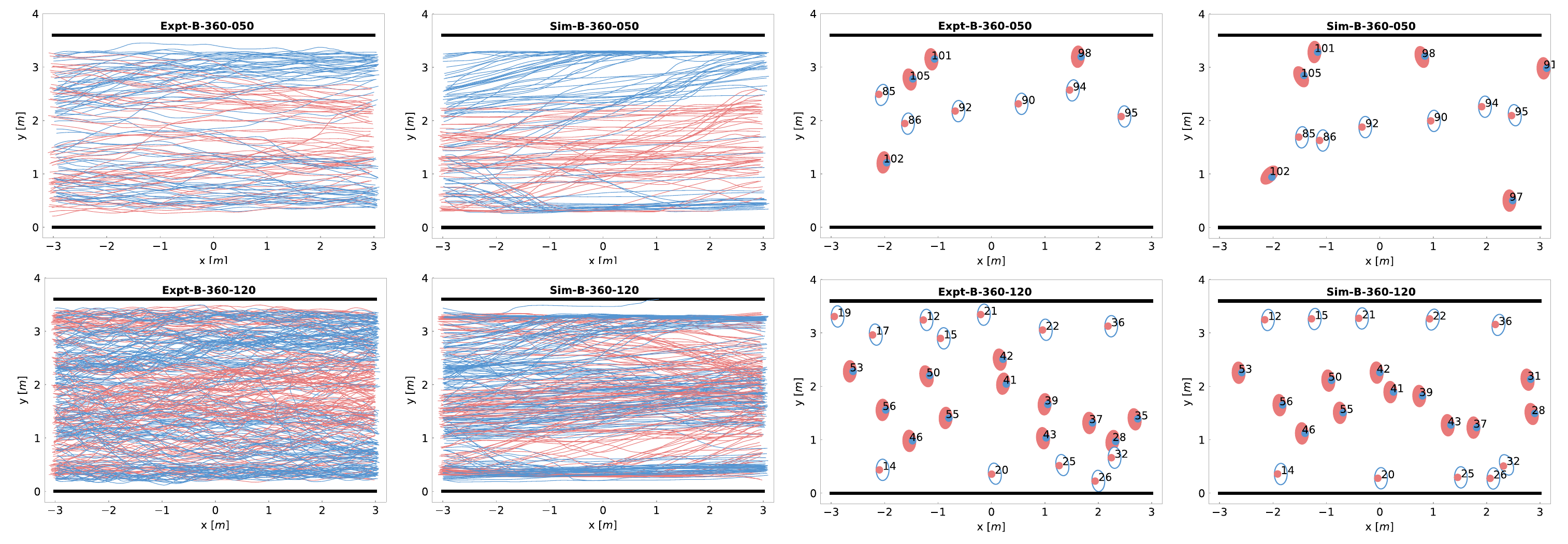}
    \caption{Pedestrian trajectories and time-aligned snapshot distributions from controlled experiments and simulations for scenarios B-360-050 and B-360-120.}
    \label{fig:traj}
    \end{figure*}

Fig. \ref{fig:traj} presents the trajectories and pedestrian snapshot distributions from controlled experiments and simulations. It can be observed that our model effectively simulates bidirectional flow, and lane formation is evident from the pedestrian snapshot distributions.
\subsubsection{Lane formation}
Lane formation, as a key self-organization phenomenon in bidirectional flow, is frequently used to validate a model. Several methods have been proposed to measure lane formation, either qualitatively or quantitatively, including velocity profiles \cite{Zhang_2012}, cluster analysis \cite{10.1007/3-540-28091-X_36}, order parameter \cite{PhysRevE.75.051402, PhysRevE.85.066128, PhysRevE.94.032304}, and lane entropy \cite{XIE2022105875}. To quantify lane formation in the bidirectional flow, we adopt the order parameter method \cite{PhysRevE.75.051402, PhysRevE.85.066128, PhysRevE.94.032304} due to its strong quantitative capability.
\par
First, we divide the corridor into rows of equal width. The set of the rows can be defined as $\mathcal{K} = \{1, 2, \cdots, k, \cdots, N_k\}$, where $N_k$ is the number of the rows, and $k$ represents the row index. The width of each row is set to 0.3 m, smaller than the pedestrian diameter \cite{PhysRevE.85.066128}. Thus, $N_k = 12$ ($3.6 / 0.3 = 12$), as the corridor width is 3.6 m. For each row $k$, the local order parameter $\Omega_k$ is computed as
\begin{equation}
    \Omega_k = \left( \frac{n_k^L - n_k^R}{n_k^L + n_k^R} \right)^2
\end{equation}
where \( n_k^L \) and \( n_k^R \) denote the numbers of left-moving and right-moving pedestrians in the \( k \)-th row, respectively. The global order parameter \(\Omega \) is then given by 

\begin{equation}
    \Omega = \frac{1}{N_k} \sum_{k=1}^{N_k} \Omega_k
\end{equation}
The order parameter ranges between 0 and 1. For bidirectional flow, the closer its value is to 1, the higher the proportion of pedestrians moving in the same direction within each row, i.e., the more pronounced the lane formation. Table \ref{tab:order_parameter} presents the order parameters calculated from the controlled experiments and simulations. It can be observed that the order parameters of the simulations are very close to those obtained from the controlled experiments, and both are close to 1, indicating that our model can effectively reproduce lane formation.

\begin{table}[htb]
    \centering
    % \tiny
    \fontsize{5}{6}\selectfont 
    \caption{Order parameters from the controlled experiments and simulations.}
    \resizebox{0.6\textwidth}{!}{
    \begin{tabular}{ccccc}
      \toprule %Add a bold line to the header
       & B-360-050 & B-360-075 & B-360-090 & B-360-120\\
      \hline
      Expt & 0.96 & 0.92 & 0.89 & 0.88 \\
      Sim & 0.96 & 0.94 & 0.88 & 0.92 \\
    \bottomrule
    \end{tabular}}
    \label{tab:order_parameter}
\end{table}

\subsubsection{Collision rate}
We calculate the collision rates between pedestrians moving in the same direction and those moving in opposite directions. Based on human body dimensions, each pedestrian here is modeled as an ellipse with a shoulder width of 0.4m and a chest depth of 0.24m. A collision is defined as occurring when two such ellipses overlap. Table \ref{tab:collision_rate} presents the collision rates obtained from controlled experiments and simulations across different model variants. Specifically, the AVBW-model refers to a variant incorporating four generator loss components: adversarial loss $\mathcal{L}_{adv}^{(i,t)}$, velocity L2 loss $\mathcal{L}_{vel}^{(i,t)}$, backward loss $\mathcal{L}_{bac}^{(i,t)}$ and wall loss $\mathcal{L}_{wal}^{(i,t)}$. The AVBWC-model extends the AVBW-model by additionally including the collision loss $\mathcal{L}_{col}^{(i,t)}$. It can be observed that the AVBW-model, which does not incorporate the collision loss $\mathcal{L}_{col}^{(i,t)}$ or the distance loss $\mathcal{L}_{dis}^{(i,t)}$, exhibits high collision rates for both same-direction and pposite-direction pedestrians. The CPGAN model demonstrates collision rates for opposite-direction pedestrians that are comparable to those obtained from controlled experiments, indicating that our proposed collision loss effectively addresses collisions between opposite-direction pedestrians. Although the CPGAN model shows higher collision rates for same-direction pedestrians compared to the experimental results, a comparison with the AVBWC-model reveals that the addition of the distance loss $\mathcal{L}_{dis}^{(i,t)}$ reduces same-direction pedestrian collisions by 36\% overall. This demonstrates a certain level of effectiveness for the distance loss. Comparing the AVBW-model and the CPGAN model, it is evident that the collision loss significantly reduces opposite-direction pedestrian collisions (by 96\% overall), while the distance loss effectively decreases same-direction pedestrian collisions (by 62\% overall).

\begin{table*}[htb]
    \centering
    \footnotesize
    \caption{Comparison of collision rates ($\%$) obtained from controlled experiments and simulations for different model variants.}
        \resizebox{1.02\textwidth}{!}{
    \begin{tabular}{cccccccccccc}
      \toprule %Add a bold line to the 
    
     & \multicolumn{2}{c}{Experiment} & & \multicolumn{2}{c}{AVBW-model} & & \multicolumn{2}{c}{AVBWC-model} & & \multicolumn{2}{c}{CPGAN}\\
                         \cline{2-3} \cline{5-6} \cline{8-9} \cline{11-12}
     & opposite & same & & opposite & same  & & opposite & same  & & opposite & same \\
    \hline
    
    B-360-050 & 0.11 &	0.19 & & 4.73 & 7.18 & & 0.25 & 1.50 & & 0.05 & 1.21 \\
    B-360-075 & 0.19 &	0.27 & & 6.61 & 8.95 & & 0.43 & 4.06 & & 0.49 & 3.46 \\
    B-360-090 & 0.50 &	0.29 & & 10.65 & 14.40 & & 0.56	& 10.26	& & 0.48 & 9.77 \\
    B-360-120 & 0.36 &	1.57 & & 12.16 & 23.03 & & 0.42	& 14.22 & & 0.34 & 6.57\\
    overall   & 0.34 &  0.93 & & 9.87 &	16.27 & & 0.43 & 9.56 & & 0.36 & 6.13 \\

    \bottomrule
    \end{tabular}    
    }
    \label{tab:collision_rate}
    \end{table*}  

\subsubsection{$N$-$t$ curves}
\begin{figure*}[h]
    \centering
    \includegraphics[width=0.98\textwidth]{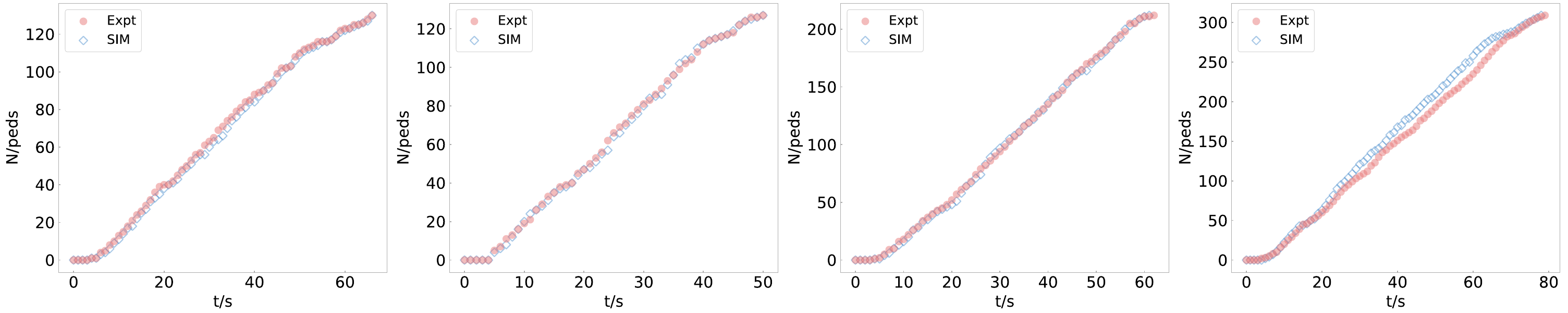}
    \caption{$N$-$t$ curves from the controlled experiments and simulations.}
    \label{fig:N-t}
    \end{figure*}

The $N$-$t$ curve shows the temporal evolution of the cumulative number of exited pedestrians. Fig.~\ref{fig:N-t} presents the $N$-$t$ curves from the controlled experiments and simulations. It can be observed that our model effectively reproduces the $N$-$t$ dynamics, as the simulated curves are in good agreement with those from the controlled experiments.

\section{Conclusion}
This paper proposes a novel CPGAN model that incorporating pedestrian collision theory into the loss function to reduce collisions with oncoming pedestrians in data-driven crowd simulation. Additionally, a Voronoi-based motion feature extraction approach is introduced. We evaluate the proposed model in bidirectional flow scenarios. Multiple metrics are utilized to assess model performance, including pedestrian trajectories and spatial distributions, lane formation, collision rates and $N$-$t$ curves. Results show that CPGAN effectively simulates bidirectional flow, reproducing lane formation and $N$-$t$ curves. The proposed collision loss significantly mitigates opposite-direction pedestrian collisions, effectively reducing the collision rate to levels comparable to those observed in controlled experiments. These outcomes represent an effective integration of pedestrian dynamics theory into the loss function framework for data-driven crowd simulation. Our approach may provide valuable insights for future research endeavors focused on incorporating various pedestrian dynamics theories into loss functions, potentially leading to further improvements in the realism of data-driven crowd simulation models.
\par

\appendix
\section{Expected minimum distance calculation } \label{ap:Minimum_distance_calculation}
To ensure the differentiability of the collision loss $\mathcal{L}_{col}^{(i,t)}$, we adopt a time-discretization approach to compute $d^{(i,j,t)}_{\min}$. Specifically, starting from current time $\tau_0$, we sample $N_{samples}$ time points over the interval $[\tau_0, \tau_{max}]$ and compute the distance between pedestrians $i$ and $j$ at each sampled time. we set $\tau_{max} = \tau_0 + 5$ (in seconds) and $N_{samples} = 1000$ to secure sufficient sampling while balancing computational cost. Based on Assumption \ref{ass:main}, the computational steps for $d^{(i,j,t)}_{\min}$ are presented in Algorithm \ref{alg:min_dis}.

\begin{algorithm}
\caption{Expected Minimum Distance Calculation via Time Discretization}
\label{alg:min_dis}
\begin{algorithmic}[1]
\Require 
    State of pedestrian $i$: position $\mathbf{p}^{(i,t)} = [p^{(i,t)}_x, p^{(i,t)}_y]$, velocity $\mathbf{v}^{(i,t)} = [v^{(i,t)}_x, v^{(i,t)}_y]$, $y$-acceleration $acc^{(i,t)}_y$
    State of pedestrian $j$: position $\mathbf{p}^{(j,t)} = [p^{(j,t)}_x, p^{(j,t)}_y]$, velocity $\mathbf{v}^{(j,t)} = [v^{(j,t)}_x, v^{(j,t)}_y]$
    Sampling parameters: max time $\tau_{max}$, number of samples $N_{samples}$
\Ensure 
    Minimum distance $d^{(i,j,t)}_{\min}$

\State \textbf{Calculate relative states:}
\State $\Delta p_x \gets p^{(i,t)}_x - p^{(j,t)}_x$
\State $\Delta p_y \gets p^{(i,t)}_y - p^{(j,t)}_y$
\State $\Delta v_x \gets v^{(i,t)}_x - v^{(j,t)}_x$
\State $\Delta v_y \gets v^{(i,t)}_y - v^{(j,t)}_y$
\State $\Delta acc_y \gets acc^{(i,t)}_y$

\State \textbf{Compute coefficients for squared distance polynomial $D^2(\tau) = c_4 \cdot (\tau-\tau_0)^4 + c_3 \cdot (\tau-\tau_0)^3 + c_2 \cdot (\tau-\tau_0)^2 + c_1 \cdot (\tau-\tau_0) + c_0$:}
\State $c_4 \gets 0.25 \cdot \Delta {acc_y}^2$
\State $c_3 \gets \Delta v_y \cdot \Delta acc_y$
\State $c_2 \gets {\Delta v_x}^2 + {\Delta v_y}^2 + \Delta p_y \cdot \Delta acc_y$
\State $c_1 \gets 2 \cdot (\Delta p_x \cdot \Delta v_x + \Delta p_y \cdot \Delta v_y)$
\State $c_0 \gets {\Delta p_x}^2 + {\Delta p_y}^2$

\State \textbf{Generate discrete time samples:}
\State $\tau_{samples} \gets \text{Linspace}(\tau_0, \tau_{max}, N_{samples})$

\State \textbf{Calculate squared distances for all samples:}
\State $D^2_{samples} \gets c_4 \cdot (\tau_{samples} - \tau_0)^4 + c_3 \cdot (\tau_{samples} - \tau_0)^3 + c_2 \cdot (\tau_{samples} - \tau_0)^2 + c_1 \cdot (\tau_{samples} - \tau_0) + c_0$
\State \textbf{Extract minimum distance:}
\State $d^2_{min} \gets \min(D^2_{samples})$
\State $d^{(i,j,t)}_{\min} \gets \sqrt{d^2_{min}}$

\Return $d^{(i,j,t)}_{\min}$
\end{algorithmic}
\end{algorithm}

\section*{Acknowledgments}
The work described in this paper was fully supported by a grant from CityU (Project No. DON\_RMG  9229030).

\bibliographystyle{elsarticle-num} 
\bibliography{ref}
\end{document}